\documentclass{article}

\usepackage{times}
\usepackage{graphicx} 
\usepackage{subfigure} 

\usepackage{natbib}

\usepackage{algorithm}
\usepackage{algorithmic}

\PassOptionsToPackage{hyphens}{url}
\usepackage{hyperref}
\usepackage[hyphenbreaks]{breakurl}

\usepackage[accepted]{icml2017}

\icmltitlerunning{Interactive Passive Automata Learning}

\begin{document} 

\twocolumn[
\icmltitle{Human in the Loop: Interactive Passive Automata Learning\\ via Evidence-Driven State-Merging Algorithms}

\begin{icmlauthorlist}
\icmlauthor{Christian A. Hammerschmidt}{lux}
\icmlauthor{Radu State}{lux}
\icmlauthor{Sicco Verwer}{delft}
\end{icmlauthorlist}

\icmlaffiliation{lux}{University of Luxembourg}
\icmlaffiliation{delft}{Delft Technical University}

\icmlcorrespondingauthor{Christian A. Hammerschmidt}{christian.hammerschmidt@uni.lu}

\icmlkeywords{automata learning, finite state machines, machine learning, ICML}

\vskip 0.3in
]

\begin{NoHyper}
\printAffiliationsAndNotice{}
\end{NoHyper}

\begin{abstract} 
We present an interactive version of an evidence-driven state-merging (EDSM) algorithm for learning variants of finite state automata. 
Learning these automata often amounts to recovering or reverse engineering the model generating the data despite noisy, incomplete, or imperfectly sampled data sources rather than optimizing a purely numeric target function.\newline{}
Domain expertise and human knowledge about the target domain can guide this process, and typically is captured in parameter settings.
Often, domain expertise is subconscious and not expressed explicitly. 
Directly interacting with the learning algorithm makes it easier to utilize this knowledge effectively.
\end{abstract} 

\section{Introduction}
\label{sec:intro}

Automata, and finite state machines in particular, have long been a formal model used in computer science to designing software and analyzing computer systems themselves~\cite{lee_principles_1996}.
Our knowledge about automata models comes from different source: 
Firstly, we have studied formal systems in detail and understand them as algebraic systems we can pose decision problems about.
Secondly, we have experience lots experience applying them as a model language.

In learning, these tasks are done inversely to recover, reverse-engineer, or understand an unknown data source, typically a communication protocol or software controller. 
The automata models themselves are also seen as suitable to gain epistemic insight into the system of interest \cite{hammerschmidt_interpreting_2016,vaandrager_model_2017} for humans beyond learning models that minimize a given error function, or are used in model checkers for formal analysis.

And while we understand the theory of learning automata fairly well, e.g.\ via learnability results and convergence guarantees, there is a disconnect between the theory of learning algorithms on one side, and domain experts using automata as a modeling language on the other side.
How can we bring the two sides together?
We propose an interactive version of a learning algorithm with immediate graphical output at each step.
This process brings reverse engineering closer to an iterative design process while guiding the process using state-of-the-art heuristics and algorithms.
In Section \ref{sec:preliminaries}, we briefly introduce finite state machines as automata models and outline learning approaches.
In \ref{sec:interactive}, we outline our algorithm and provide a small example outlining how our modification is useful.
We conclude in Section \ref{sec:conclusion} with an outlook and future work.

\section{Preliminaries}
\label{sec:preliminaries}

\subsection{Automata Models and Finite Automata}
\label{sec:sub:automata}

The models we consider are variants of deterministic finite state automata, or finite state machines. 
We provide a conceptual introduction here, and refer to literature for a thorough introduction \citep{hopcroft_introduction_2013}.
An automaton consists of a set of states, connected by transitions labeled over an alphabet. 
There is a special start-state where all computations begin, and and a set of final states where computation ends.
It is said to accept a word (string) over the alphabet in a computation if there exists a path of transitions from a predefined start state to one of the predefined final states, using transitions labeled with the letters of the word. 
Automata are called deterministic when there exists exactly one such path for every possible string. 
Probabilistic automata include probability values on transitions and compute word probabilities using the product of these values along a path, similar to hidden Markov models, rather than having final states. 
A Mealy machine is a deterministic finite automata that produces an output at each step.

\subsection{Learning Approaches}
\label{sec:sub:learningapproaches}

Learning automata models, and in particular finite state machines, has long been of interest in the field of grammar induction. While the problem can be solved exactly (see~\citep{heule_software_2012}), it is NP-complete. There are two main classes of learning approaches: \emph{active learning} and \emph{passive learning} approaches.

In an active setting, the active learner is given the opportunity to interact with an oracle (or teacher) to ask questions about her current hypothesis of the model.
Depending on the type of query and answers given by the oracle, different learnability results and algorithms can be obtained \cite{heinz_topics_2016}.
The $L^*$ algorithm is a successful active learner~\cite{angluin_learning_1987}.

In contrast, the learner in a passive setting has no oracle to interact with.
Rather, he is given a fixed set of training data to learn from.
In learning stochastic automata, methods based on the methods of moments as well as spectral learning algorithms have been studied~\cite{balle_spectral_2014}. In \citet{dupont_qsm_2008}, the authors combine active and passive approaches by asking a user membership queries during a state-merging process.

State-merging algorithms have been very successful in learning probabilistic as well as non-probabilistic automata, and trace back to~\citet{oncina_identifying_1992}. 
Recent state-of-the-art algorithms base on insights from \citet{lang_evidence_1998,verwer_pautomac:_2014}. 
We explain the generic algorithm and our modification in the next section.
%

\section{State-Merging and Interactive State-Merging}
\label{sec:interactive}

The starting point for state-merging algorithms is the construction of a tree-shaped automaton from a set of words, the input sample. This initial automaton is called augmented prefix tree acceptor (APTA). It contains all sequences from the input sample, with each sequence element as a directed labeled transition, and states are added to the beginning, end, and between each symbol of the sequence. Two samples share a path in the tree if they share a prefix. The state-merging algorithm reduces the size of the automaton iteratively by merging pairs of states in the model, and forcing the result to be deterministic. The choice of the pairs, and the evaluation of a merge is made heuristically: Each possible merge is evaluated and scored, and the highest scoring merge is executed. This process is repeated and stops if no merges with high scores are possible. These merges generalize the model beyond the samples from the training set: the starting prefix tree is already an acyclic finite automaton. It has a finite set of computations, accepting all words from the training set. Merges can make the resulting automaton cyclic. Automata with cycles accept an infinite set of words. State-merging algorithms generalize by identifying repetitive patterns in the input sample and creating appropriate cycles via merges. 
Intuitively, the heuristic tries to accomplish this generalization by identifying pairs of states that have similar future behaviors. In a probabilistic setting, this  similarity might be measured by similarity of the empirical probability distributions over the outgoing transition labels. In grammar inference, heuristics rely on occurrence information and the identity of symbols, or use global model selection criteria to calculate merge scores. 
Finally, the state-merging process as well as the heuristic are controlled by a number of parameters.
The general outline of the algorithm is the same as in Algorithm \ref{alg:interactive}, although the user prompt needs to be replaced with execution of the highest scoring merge.

\subsection{An Interactive Algorithm}
\label{sec:sub:interactive}

Algorithm \ref{alg:interactive} shows the modified interactive algorithm: Instead of automatically executing the best merge, the graphical visualization of the current model is displayed and the list of possible merges is displayed. 
The user can choose which of the possible merges to execute by typing its number in the sorted list of possible merges, to backtrack by one step using \texttt{UNDO}, to restart using \texttt{RESTART}, or to automatically execute the next $n$ best merges using \texttt{LEAP} $n$. 
Moreover, she can change some global parameter of the algorithm using \texttt{SET} $param$ and see the impact on the proposed merges. Lastly, she can add additional consistency constraints or force merges using the \texttt{INSERT} and \texttt{FORCE} commands.
The order of the proposed merges follows the machine learning heuristic. Choosing the first/top merge prosed will lead to the same solution as a single batch-run of the algorithm.
The source code for our C++ implementation (as part of out \emph{flexfringe} tool \citep{verwer_flexfringe:_2017}) is available in our repository\footnote{\url{https://bitbucket.org/chrshmmmr/dfasat/src/?at=development}}.

In our implementation, the algorithm keeps a stack of executed merges and a list of currently possible merges to store the current state of the algorithm. Based on these data structures, the user has full control over the state-merging process. Additionally, we output the visualization for the last two steps for visual inspection.

\subsection{Example Use Case}

\begin{figure}[ht]
\begin{center}
\centerline{\includegraphics[width=\columnwidth]{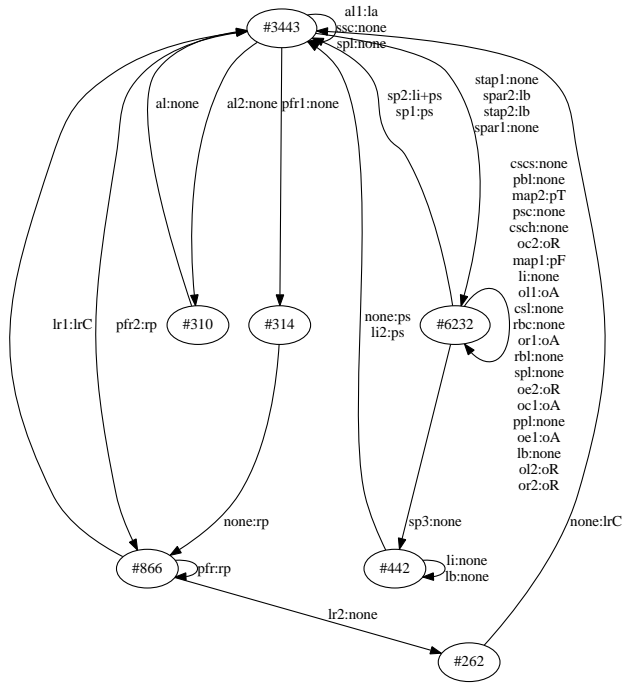}}
\caption{The original and fully recovered model of our example use case. State \#3443 is the start state. The transitions are annotated with \emph{input:output} symbols.}
\label{fig:original}
\end{center}
\vskip -0.2in
\end{figure}

We illustrate the benefits of the interactive mode in a small example: The task is to recover the Mealy machine in Figure \ref{fig:original} from traces sampled from it.\footnote{For the batch version of our algorithm, this example is provided in an Jupyter notebook environment, see\small{ \url{http://automatonlearning.net/2016/11/04/a-passive-automata-learning-tutorial-with-dfasat/}}}

\begin{figure}[ht]
\begin{center}
\centerline{\includegraphics[width=0.66\columnwidth]{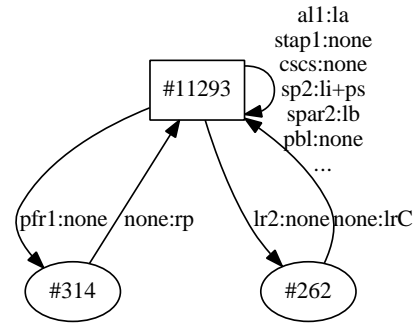}}
\caption{State \#11293 is the start state. The transitions are annotated with \emph{input:output} symbols. The $\ldots$ indicate 30 more input-output pairs on the transition. The model overgeneralizes  by merging states together despite low evidence of similarity.}
\label{fig:failure}
\end{center}
\vskip -0.2in
\end{figure}

Running an Mealy machine heuristic on 1000 sequences sampled from the model in batch mode yields a wrong model, depicted in Figure \ref{fig:failure}\footnote{To reproduce, run  \emph{./flexfringe --heuristic-name mealy --data-name mealy\_data --sinkson 1 --satdfabound 2000 data/original/traces1000.txt} on the data in the repository.}. Our state-merging algorithm outputs the following sequence of merges and extends:
\small{
\begin{verbatim}
m147 m182 m207 m10 m58 m79 m98 m47 m69 m62 
m54 m57 m69 m71 m70 m83 m81 m97 m104 m117 
m144 m158 m181 m221 m254 m305 m343 m413 
m539 m119 x162 m264 m259 m0 m1036 m0 x61 
m343 m42 m695 m780
\end{verbatim}
}

There are two types of operations in the algorithm: merges and extensions. The letter $m$, e.g.\ $m147$, indicates operation executed was a merge (e.g.\ with score of $147$). The letter $x$ indicates and extension, e.g.\ $x169$ indicates that a blue state occurring $169$ times was inconsistent with all red states, and no merge could be executed. It therefore is color red. Better evidence, indicated by higher scores and higher occurrence counts, typically leads to better models.
Two merges in the sequence above have an evidence of 0. This indicates that despite the lack of evidence for the inconsistency of the merge, there is also little evidence for the merge to be a good choice. Such merges can lead to over-generalization.

A common strategy to deal with this situation, which can occur e.g.\ due to the lack of sufficient training data or under-sampling specific paths in the model, is to adjust parameters of the state-merger or the heuristic by setting a lower bound on the required merge score and re-running the batch algorithm to resolve further issues\footnote{To reproduce the model in batch mode, run \emph{./flexfringe --heuristic-name mealy --data-name mealy\_data --state\_count 15 --symbol\_count 5 --sinkson 1 --satdfabound 2000 --lowerbound 50 data/original/traces1000.txt} on the data in the repository.}. But this is not the only possible parameter to adjust: Other strategies involve changing the relevancy threshold that ignore traces with low counts (which often can indicate the presence of noise in the collected data).

In our interactive version, a domain expert can immediately react when a merge with lower score is proposed (even without a visual inspection): She can choose to change the thresholds (commands \texttt{SET} $state\_count$ or \texttt{SET} $symbol\_count$), but they do not change the proposed merges. The expert can decide to exclude the merge by setting a lower bound (\texttt{SET} $lower\_bound$ 10), and possible use to \texttt{restart} and \texttt{leap} back to the position of the merge in question.
The resulting model, after executing all following proposed merges is the correct model (as depicted in Figure \ref{fig:original})\footnote{To reproduce, run \emph{./start.sh interactive-mealy.ini data/original/traces1000.txt} from the repository, execute merges until the first $m0$, \texttt{set} $lower\_bound$ $10$, \texttt{restart}, \texttt{leap} $35$, and execute all merges.}.

In contrast, running the algorithm in batch mode would have required to re-run the complete algorithm multiple times, and comparing the sequence of merges done without the ability to draw up a list of alternative merges for each step.

\section{Conclusion \& Future Work}
\label{sec:conclusion}

Our experience indicates that a hybrid mode between manual and fully automatic decision is most useful:
Especially in cases of large training files (and therefore large initial models which are hard to use visually), automating all decisions until a specific criterion is met (e.g.\ the best proposed score is below a given threshold, or $n$ merges are executed) takes some burden off the human domain expert while still allowing intervention when necessary. 
By following the suggestions of the heuristic, the expert can retain learnability guarantees from batch learning algorithms.
We have not yet conducted a (formal) analysis to quantify the benefits of interactivity in terms of convergence speed. 
In \citet{lambeau_state-merging_2008}, the authors experimentally showcase the positive effect of mandatory merge constraints. Similar information, and thus benefits, are provided by the interaction proposed here.
Domain expert can implicitly act as a teacher or oracle active learning,  bringing together elements from both active and passive learning approaches to automata.

\section*{Acknowledgements} 
 I would like to thank my collages for the valuable discussions and the feedback. 

\begin{algorithm}[ht]
\caption{Interactive Evidence-Driven State-Merging \label{alg:interactive}}
\begin{algorithmic}
\REQUIRE a set of input words $S$
\ENSURE $\mathit{A}$ is a DFA that is consistent with $S$ 
\STATE $\mathit{A}$ = {\sf apta}$(S)$ \COMMENT{construct the APTA $\mathit{A}$}
\STATE $M$ = empty\_stack() \COMMENT{to store merge history}
\STATE $R = \{q_0\}$ \COMMENT{color the start state of $\mathit{A}$ red}
\STATE $B = \{q \in Q \setminus R \mid \exists \left<q_0, q, l\right> \in T\}$ \COMMENT{color all its children blue}
\WHILE[while $\mathit{A}$ contains blue states]{ $B \not= \emptyset$ }
\IF[if a blue is state inconsistent with all red states]{ $\exists b \in B$ s.t. $\forall r \in R$ holds $merge(\mathit{A},r,b) =$ {\sc false}}
\STATE $R := R \cup \{b\}$ \hfill // color $b$ red
\STATE $B := B \cup \{q \in Q \setminus R \mid \exists \left<b, q, l\right> \in T\}$ \COMMENT{color all its children blue}
\ELSE
\STATE L := empty\_list()
\FORALL[forall red-blue pairs of states] {$b \in B$ and $r \in R$ }
\STATE compute $\textsf{evidence}(\mathit{A},q,q')$ of $merge(\mathit{A},r,b)$ \COMMENT{find merge pairs and scores}
\STATE l.add($\textsf{evidence}(\mathit{A},q,q')$)
\ENDFOR
\STATE \texttt{p := prompt user}
\IF[handle(p)]{p not a merge}
\STATE \COMMENT{undo, restart, leap, or set parameters}
\STATE $continue$
\ENDIF
\STATE $\mathit{A} := merge(\mathit{A},r,b)$ as user chose \COMMENT{perform the chosen merge}
\STATE M.push$(r,b)$
\STATE let $q''$ be resulting state
\STATE $R := R \cup \{q''\}$ \COMMENT{color the resulting state red}
\STATE $R := R \setminus \{r\}$ \COMMENT{uncolor the merged red state}
\STATE $B := \{q \in Q \setminus R \mid \exists r \in R \wedge \left<r, q, l\right> \in T\}$ \COMMENT{recompute the set of blue states}
\ENDIF
\ENDWHILE
\OUTPUT $\mathit{A}$
\end{algorithmic}
\end{algorithm}

\bibliography{Zotero}

\begin{thebibliography}{14}
\providecommand{\natexlab}[1]{#1}
\providecommand{\url}[1]{\texttt{#1}}
\expandafter\ifx\csname urlstyle\endcsname\relax
  \providecommand{\doi}[1]{doi: #1}\else
  \providecommand{\doi}{doi: \begingroup \urlstyle{rm}\Url}\fi

\bibitem[Angluin(1987)]{angluin_learning_1987}
Angluin, Dana.
\newblock Learning regular sets from queries and counterexamples.
\newblock \emph{Information and computation}, 75\penalty0 (2):\penalty0
  87--106, 1987.

\bibitem[Balle et~al.(2014)Balle, Carreras, Luque, and
  Quattoni]{balle_spectral_2014}
Balle, Borja, Carreras, Xavier, Luque, Franco~M., and Quattoni, Ariadna.
\newblock Spectral learning of weighted automata: {A} forward-backward
  perspective.
\newblock \emph{Machine Learning}, 96\penalty0 (1-2):\penalty0 33--63, July
  2014.
\newblock ISSN 0885-6125, 1573-0565.
\newblock \doi{10.1007/s10994-013-5416-x}.

\bibitem[Dupont et~al.(2008)Dupont, Lambeau, Damas, and
  Lamsweerde]{dupont_qsm_2008}
Dupont, Pierre, Lambeau, Bernard, Damas, Christophe, and Lamsweerde, Axel~van.
\newblock {THE} {QSM} {ALGORITHM} {AND} {ITS} {APPLICATION} {TO} {SOFTWARE}
  {BEHAVIOR} {MODEL} {INDUCTION}.
\newblock \emph{Applied Artificial Intelligence}, 22\penalty0 (1-2):\penalty0
  77--115, February 2008.
\newblock ISSN 0883-9514, 1087-6545.
\newblock \doi{10.1080/08839510701853200}.

\bibitem[Hammerschmidt et~al.(2016)Hammerschmidt, Lin, Verwer, and
  State]{hammerschmidt_interpreting_2016}
Hammerschmidt, Christian~Albert, Lin, Qin, Verwer, Sicco, and State, Radu.
\newblock Interpreting {Finite} {Automata} for {Sequential} {Data}.
\newblock \emph{arXiv:1611.07100 [cs, stat]}, November 2016.
\newblock arXiv: 1611.07100.

\bibitem[Heinz \& Sempere(2016)Heinz and Sempere]{heinz_topics_2016}
Heinz, Jeffrey and Sempere, José~M.
\newblock \emph{Topics in {Grammatical} {Inference}}.
\newblock Springer, 2016.

\bibitem[Heule \& Verwer(2012)Heule and Verwer]{heule_software_2012}
Heule, Marijn J.~H. and Verwer, Sicco.
\newblock Software model synthesis using satisfiability solvers.
\newblock \emph{Empirical Software Engineering}, 18\penalty0 (4):\penalty0
  825--856, August 2012.
\newblock ISSN 1382-3256, 1573-7616.
\newblock \doi{10.1007/s10664-012-9222-z}.

\bibitem[Hopcroft et~al.(2013)Hopcroft, Motwani, and
  Ullman]{hopcroft_introduction_2013}
Hopcroft, John~E., Motwani, Rajeev, and Ullman, Jeffrey~D.
\newblock \emph{Introduction to {Automata} {Theory}, {Languages}, and
  {Computation}}.
\newblock Pearson, Harlow, Essex, pearson new international edition edition,
  November 2013.
\newblock ISBN 978-1-292-03905-3.

\bibitem[Lambeau et~al.(2008)Lambeau, Damas, and
  Dupont]{lambeau_state-merging_2008}
Lambeau, Bernard, Damas, Christophe, and Dupont, Pierre.
\newblock State-merging {DFA} induction algorithms with mandatory merge
  constraints.
\newblock \emph{Grammatical Inference: Algorithms and Applications}, pp.\
  139--153, 2008.

\bibitem[Lang(1998)]{lang_evidence_1998}
Lang, K.
\newblock Evidence driven state merging with search.
\newblock 1998.

\bibitem[Lee \& Yannakakis(1996)Lee and Yannakakis]{lee_principles_1996}
Lee, D. and Yannakakis, M.
\newblock Principles and methods of testing finite state machines-a survey.
\newblock \emph{Proceedings of the IEEE}, 84\penalty0 (8):\penalty0 1090--1123,
  August 1996.
\newblock ISSN 0018-9219.
\newblock \doi{10.1109/5.533956}.

\bibitem[Oncina \& Garcia(1992)Oncina and Garcia]{oncina_identifying_1992}
Oncina, Jose and Garcia, Pedro.
\newblock Identifying {Regular} {Languages} {In} {Polynomial} {Time}.
\newblock In \emph{Advances in ...}, pp.\  99--108. World Scientific, 1992.

\bibitem[Vaandrager(2017)]{vaandrager_model_2017}
Vaandrager, Frits.
\newblock Model {Learning}.
\newblock \emph{Commun. ACM}, 60\penalty0 (2):\penalty0 86--95, January 2017.
\newblock ISSN 0001-0782.
\newblock \doi{10.1145/2967606}.

\bibitem[Verwer et~al.(2014)Verwer, Eyraud, and
  De~La~Higuera]{verwer_pautomac:_2014}
Verwer, Sicco, Eyraud, Rémi, and De~La~Higuera, Colin.
\newblock {PAutomaC}: a probabilistic automata and hidden {Markov} models
  learning competition.
\newblock \emph{Machine learning}, 96\penalty0 (1-2):\penalty0 129--154, 2014.

\bibitem[Verwer \& Hammerschmidt(2017)Verwer and
  Hammerschmidt]{verwer_flexfringe:_2017}
Verwer, Sicco~E. and Hammerschmidt, Christian~A.
\newblock flexfringe: {A} {Passive} {Automaton} {Learning} {Package}.
\newblock September 2017.

\end{thebibliography}
\bibliographystyle{icml2017}

\end{document}